\documentclass{article}

\usepackage{arxiv}

\usepackage[utf8]{inputenc} 
\usepackage[T1]{fontenc}    
\usepackage{url}            
\usepackage{booktabs}       
\usepackage{amsfonts}       
\usepackage{nicefrac}       
\usepackage{microtype}      
\usepackage{xcolor}         
\usepackage{wrapfig}
\usepackage{graphicx} 
\usepackage{enumitem}
\usepackage{amssymb}
\usepackage{amsmath}
\usepackage{algorithm}
\usepackage{algpseudocode}
\usepackage[normalem]{ulem}
\useunder{\uline}{\ul}{}
\usepackage[colorlinks,linkcolor=red,anchorcolor=blue,citecolor=green]{hyperref}
\usepackage{multirow}
\usepackage{tcolorbox}
\usepackage{natbib}

\title{MeGU: Machine-Guided Unlearning with Target Feature Disentanglement}

\author{%
  Haoyu Wang$^{1}$,
  Zhuo Huang$^{2}$,
  Xiaolong Wang$^{1}$,
  Bo Han$^{3}$,
  Zhiwei Lin$^{1}$,
  Tongliang Liu$^{2}$\\
  \small{$^1$School of Automation, Beijing Institute of Technology;}
  \small{$^2$Sydney AI Centre, The University of Sydney;}\\
  \small{$^3$Department of Computer Science, Hong Kong Baptist University}
}

\date{}

\renewcommand{\shorttitle}{Machine-Guided Unlearning with Target Feature Disentanglement}

\begin{document}
\maketitle

\begin{abstract}
The growing concern over training data privacy has elevated the "Right to be Forgotten" into a critical requirement, thereby raising the demand for effective Machine Unlearning. However, existing unlearning approaches commonly suffer from a fundamental trade-off: aggressively erasing the influence of target data often degrades model utility on retained data, while conservative strategies leave residual target information intact.
In this work, the intrinsic representation properties learned during model pretraining are analyzed. It is demonstrated that semantic class concepts are entangled at the feature-pattern level, sharing associated features while preserving concept-specific discriminative components. This entanglement fundamentally limits the effectiveness of existing unlearning paradigms.
Motivated by this insight, we propose Machine-Guided Unlearning (MeGU), a novel framework that guides unlearning through concept-aware re-alignment. Specifically, Multi-modal Large Language Models (MLLMs) are leveraged to explicitly determine re-alignment directions for target samples by assigning semantically meaningful perturbing labels. To improve efficiency, inter-class conceptual similarities estimated by the MLLM are encoded into a lightweight transition matrix.
Furthermore, MeGU introduces a positive-negative feature noise pair to explicitly disentangle target concept influence. During finetuning, the negative noise suppresses target-specific feature patterns, while the positive noise reinforces remaining associated features and aligns them with perturbing concepts. This coordinated design enables selective disruption of target-specific representations while preserving shared semantic structures.
As a result, MeGU enables controlled and selective forgetting, effectively mitigating both under-unlearning and over-unlearning.
Extensive experiments across varying unlearning scenarios and diverse datasets demonstrate that MeGU consistently outperforms state-of-the-art baselines, achieving thorough target data removal while maintaining strong generalization performance on retained data.	
\end{abstract}

\section{Introduction}\label{sec:introduction}

Machine Learning (ML), based on utilization of data resources, has been extensively explored in vast fields \citep{jordan2015machine,lu2007imageclassificationsurvey,nadkarni2011NLPsurvey}. However, studies \citep{jegorova2022privacyleakagesurvey, fu2022knowledge} highlight the potential data security risks such as differential privacy \citep{dwork2014algorithmic} and adversarial vulnerabilities \citep{steinhardt2017certified}. Regulations \citep{voigt2017euGDPR, bonta2022california} have been introduced to address these concerns, where the "Right to be Forgotten" \citep{rosen2011right} is underlined to enable deletion request from data providers to protect their privacy. 
On the other hand, part of the information exposed to a model during the training phase is often time-sensitive or inherently unreliable, such as noisy labels. Eliminating the influence of such unreliable information on a deployed model without resorting to retraining constitutes another critical requirement for achieving trustworthy AI~\citep{hine2024supporting,kang2025unlearning,huang2025trustworthy,wang2024noisegpt,li2024dynamic,yuan2024early,yuan2025enhancing}.
To fulfill this target, an intuitive solution is to train a new model from scratch without target data. However, the substantial computation and time costs are simply unacceptable for practices. 
Thereby, the Machine Unlearning (MU) is proposed to tackle this problem, whose objectives are: 1) Remove the influence of specific data while maintaining the overall generalization; 2) Cost less time and computational resources than simply retraining. 

\par

Prior researchers mainly explore two types of unlearning strategies: model-centric and data-centric approaches. 
Model-centric methods directly modify model architectures or parameters through alternative training paradigms \citep{bourtoule2021SISA,yan2022ARCANE} or influence estimation \citep{sekhari2021remember,zhao2022pruning}. While data-centric approaches \citep{golatkar2020eternalFIM, shibata2021learning} utilize re-optimization \citep{golatkar2020eternalFIM} or gradient updates \citep{wu2020deltagrad,huang2026gradient} to align the unlearned model with the retrained model, called "gold model".
However, these approaches often face problems such as storage overhead and inaccurate unlearning.
Another line of works \citep{graves2021amnesiac}, such as UNSIR \citep{tarun2023fastUNSIR}, crave to leverage feature noises to enhance forgetting of target data during fine-tuning. However, empirical evidence suggests a trade-off between effectively eliminating the influence of target data and maintaining the overall generalization capability \citep{ye2021towards,huang2023harnessing,huang2025towards}.
\par

\begin{wrapfigure}{r}{0.5\textwidth}
    \centering
    \vspace{-5mm}
    \includegraphics[width=1\linewidth]{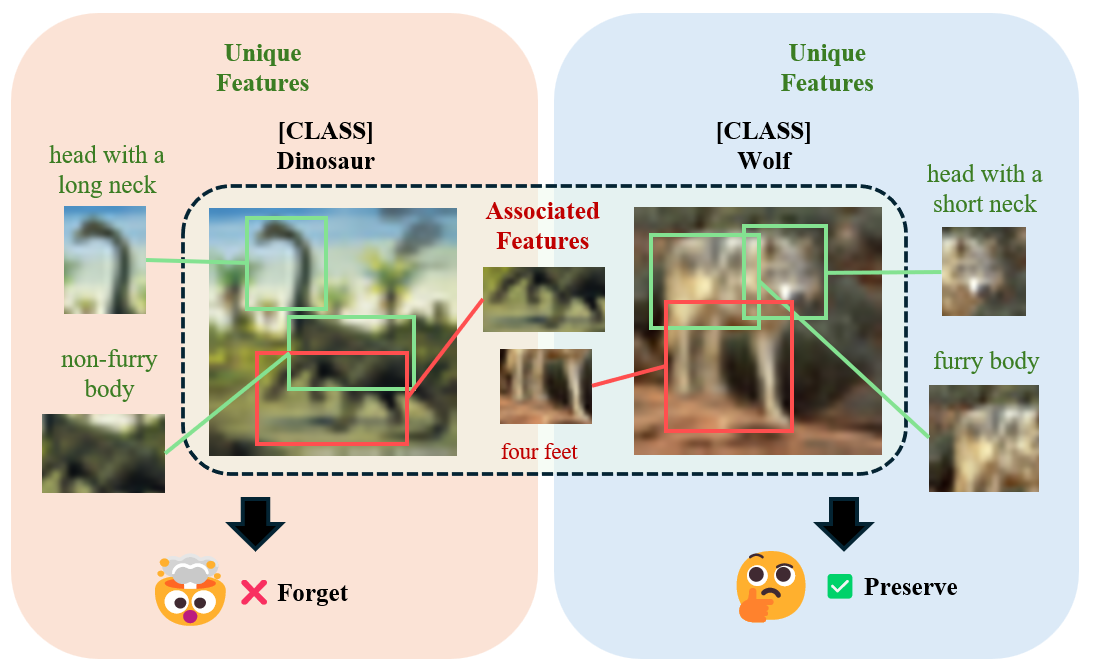}
    \vspace{-6mm}
    \caption{The entanglement among features from different concepts. Taking dinosaur and wolf as an example. They share similar features (marked as \textcolor{red}{red}) while each possesses unique features (\textcolor{green}{green}). Assume that dinosaur is the class to be forgotten. The goal of our method is to disentangle its features and forget the target concept's unique features while preserving associated features shared with the other retained concepts.}
    \vspace{-2mm}
    \label{fig:FeatureAndConcept}
\end{wrapfigure}

To address this challenge, we first investigate in a question: \textit{How does the unlearned data affect the retained model generalization?} Existing research \citep{serra2018overcoming} indicates that models acquire two levels of cognition during pretraining: \textbf{feature patterns} directly extracted from data, and \textbf{semantic concepts} representing complex relations and combinations of different feature patterns, during pretraining. Prior study \citep{chang2024class} interprets it as a mapping of features onto a higher-dimensional concepts space. 
This explains the interleaved influence that unlearned data poses on the retained model generalization, as different semantic concepts could share certain amounts of patterns, hereby named as associated features, while they each have unique features to distinguish from others. 
Figure \ref{fig:FeatureAndConcept} shows such an example. When unlearning one of the concept, the associated features between two concepts are restrained, harnessing the generalization on the retained concept. Such interplay underscores the nature of the trade-off challenge.
However, previous studies focus either on eliminating all related feature patterns of target concept \citep{tarun2023fastUNSIR} or re-aligning the target concept with other retained concepts \citep{chen2023boundary}. The complex entanglement at the feature pattern level is neglected, leading to excessive or insufficient unlearning. 
\par


In this paper, we propose Machine-Guided Unlearning (MeGU) as a framework to manipulate and enhance the unlearning process, as shown in Figure \ref{fig:framework}. Intuitively, MeGU aims to unlearn certain data by injecting misleading concepts to ensure learning a desired semantic gap from the target data. To achieve this, we leverage the guidance of zero-shot Multi-modal Large Language Models (MLLMs)~\citep{liu2024visual,huang2024machine} to generate perturbing labels for finetuning. The similarities between different concepts are based on the understanding of prompted MLLMs. We store such a similarity in a transition matrix, which facilitates efficient inference of subsequent instance-based feature understanding.
Furthermore, to address the problem of feature entanglement, we introduce the Fragment-Align strategy, which disentangles semantic concepts via a positive-negative feature noise pair~\cite{hong2024improving,hong2022semantic,chen2025semantics}. 
Specifically, the positive feature noise aims to align the target data representation with the semantics of the perturbing labels, while the negative feature noise disrupts the original representation associated with the true labels. Working together, these two complementary noises disentangle the target features from their original concepts and re-anchor them toward the perturbed concepts, enabling selective forgetting without harming overall generalization.
Through substantial studies on three unlearning tasks conducted on a range of datasets, the efficacy and efficiency of MeGU as an unlearning method are rigorously validated. Our contributions in this paper can be summarized into:
\begin{itemize}[left=1em]
    \item We investigate the intrinsic mechanisms underlying model generalization from a two-level cognitive perspective, namely feature patterns and semantic concepts. We further uncover the entangled interactions among different concepts, which give rise to the inherent trade-off challenges faced by existing unlearning approaches.
    \item We propose MeGU, a novel framework that leverages MLLM guidance to explicitly manipulate the unlearning process. Building upon this framework, we introduce the Fragment-Align strategy to disentangle the influence of target data, thereby effectively alleviating the aforementioned trade-off. Moreover, we provide both theoretical analysis and empirical evidence to systematically differentiate MeGU from representative unlearning paradigms.
    \item We conduct extensive evaluations across three distinct unlearning scenarios to rigorously assess the effectiveness of the proposed method. In addition, we perform comprehensive ablation and sensitivity studies to examine the contributions of individual components within the framework as well as the impact of key hyperparameter settings. Detailed analyses of experimental results are presented to substantiate our claims.
\end{itemize}

\section{Related works}
\subsection{Machine unlearning}
Existing Machine Unlearning (MU) approaches can be categorized into two branches: model-centric and data-centric unlearning. Model-centric unlearning endeavors to repurpose knowledge from pre-trained models by selectively modifying or filtering their components or parameters, such as Sharded, Isolated, Sliced, and Aggregated (SISA) training \citep{bourtoule2021SISA}. Researchers \citep{yan2022ARCANE, zhou2022DSMixup, brophy2021DaRE} continue to improve the performance through techniques such as data preprocessing. However, isolated training of model components will lead to generalization degradation and additional costs for initial training and storage.
Model pruning \citep{zhao2022pruning, liu2024model, wang2022federated, tanaka2020pruning, feldman2020does,stephenson2021geometry}, emerges to discard target data by selectively manipulating related crucial parameters \citep{yeom2021pruning-magitude, ma2021sanity, frankle2018lottery}. Besides, the influence functions are typically approximated to estimate the essence of parameters to target data \citep{wu2022puma, sekhari2021remember, suriyakumar2022algorithms, foster2024fastSSD, golatkar2020eternalFIM,yang2025exploring}. 
Nevertheless, the risk of over-unlearning persists, as parameters important for forgetting data may also play a significant role for retained data \citep{chang2024class}. 
Data-centric unlearning aims to adjust pre-trained models through re-optimization and gradient updates \citep{neel2021descent, cao2023fedrecover, graves2021amnesiac, shaik2024exploring, fan2023salun, yue2025what,wang2025gru}to address forgetting requests. \citep{golatkar2020eternalFIM} introduce a scrubbing function during finetuning stage to align the unlearned model with the "gold model", which is continuously refined by subsequent research \citep{golatkar2021mixed, golatkar2020forgetting, shibata2021learning, mehta2022deep, tanno2022repairing}. DeltaGrad \citep{wu2020deltagrad} and BAERASER \citep{liu2022backdoorBAERASER} employ gradient updates using cached weight information during training to unlearn target data. Despite their achievements, methods of this kind rely on strong convexity assumptions, leading to unavoidable approximation errors. 
Alternative approaches have explored the utilization of feature noise \citep{tarun2023fastUNSIR,wang2024sober} and label noise \citep{chen2023boundary} to diminish the generalization of forgetting data. Teacher-student framework \citep{chundawat2023canbadteacher, zhang2023machine, lin2023erm,huang2022they} is also employed to selectively distill knowledge, excluding forgetting data. While these methods provide intuitive solutions, they still face the inherent trade-offs problem of excessive or insufficient unlearning.
\par

\subsection{In-context Learning of Multi-modal large language models}
The emergence of Transformer \citep{vaswani2017attention} facilitates the development of Large Language Models (LLM) \citep{brown2020language, floridi2020gpt, touvron2023llama, touvron2023llama2}. 
Incorporated with Vision Transformer (ViT) \citep{dosovitskiy2020image}, LLM is equipped with efficient multi-modal capabilities, known as Multi-modal Large Language Models (MLLM) \citep{li2022blip, li2023blip, liu2024visual, dai2023instructblip, ye2023mplug}.
By integrating Vision Transformers (ViT) \citep{dosovitskiy2020image} and cross-modal alignment modules, LLMs are extended to Multi-modal Large Language Models (MLLMs), enabling joint reasoning over text and visual inputs \citep{li2022blip, li2023blip, dai2023instructblip, ye2023mplug}. Within this framework, the representations of text and images are aligned through an intermediate structure. 
In this setting, ICL allows models to perform zero-shot or few-shot task adaptation by conditioning on input demonstrations, without updating model parameters \citep{brown2020language}. Studies suggest that ICL can be interpreted as a form of implicit meta-learning or Bayesian inference over latent task structures \citep{garg2022can, xie2021explanation}.
Building on this paradigm, MMICL \citep{zhao2023mmicl} introduces a context training strategy that interleaves visual embeddings with textual tokens, significantly enhancing multimodal in-context generalization. Related works \citep{alayrac2022flamingo, liu2024visual} further indicate that multimodal ICL not only supports task adaptation, but also reveals the internal alignment and similarity structure of learned visual–semantic concepts. These properties make MLLMs a natural tool for probing conceptual relationships across classes.
In this paper, we propose to leverage the strengths of MLLMs to guide the unlearning process through a novel perturbing label assigning strategy. The in-context learning capabilities of MLLMs are utilized to estimate the similarities of feature patterns among different class concepts, which subsequently guarantees the disentanglement of influence of target data. \par

\begin{figure*}[ht]
    \centering
    \includegraphics[width=0.9\linewidth]{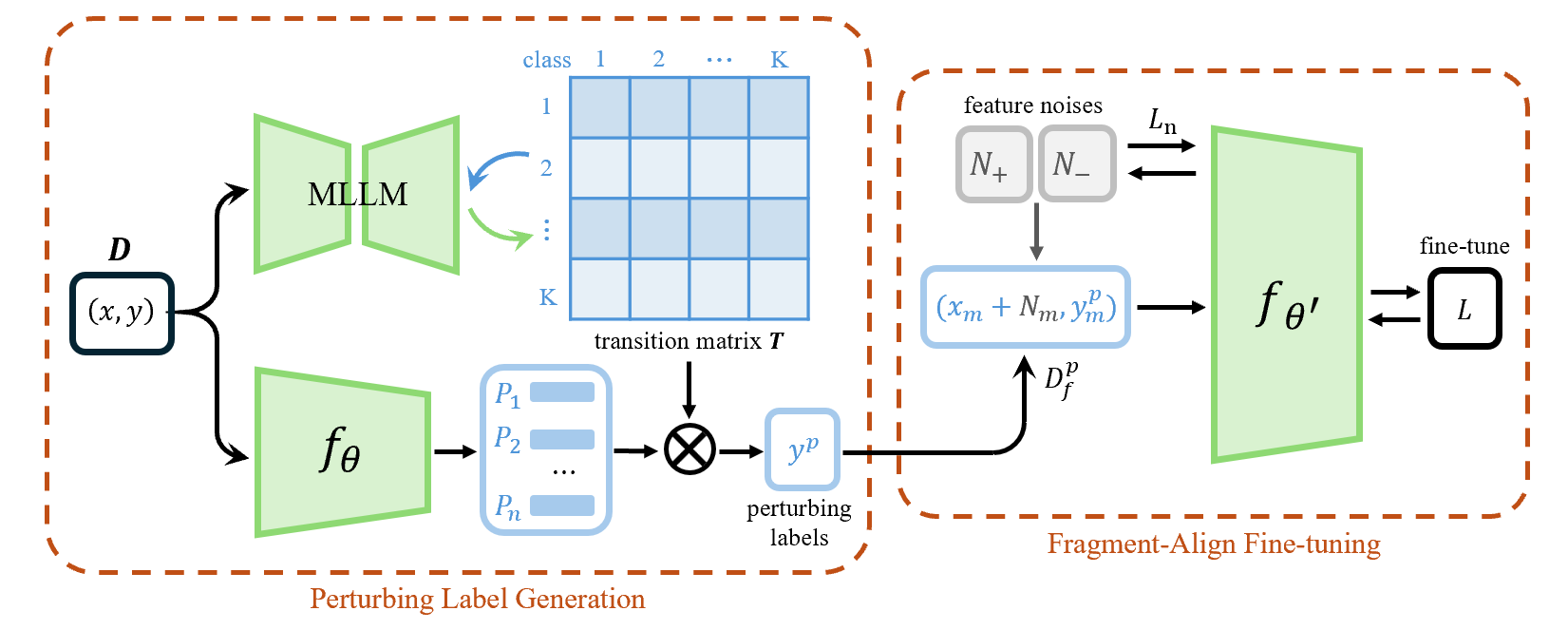}
    \vspace{-2mm}
    \caption{The proposed unlearning framework MeGU. The MLLM is employed to acquire the conceptual similarities with a small subset of the training data. Incorporated with model prediction, the perturbing labels are determined.
    The Fragment-Align strategy leverages a pair of feature noises trained from the frozen pretrained model to respectively restrain unique features of the target data while enhancing those associated with the retained data. The unlearning is achieved via aligning the target data towards perturbing labels while disentangling their influence.}
    \vspace{-2mm}
    \label{fig:framework}
\end{figure*}

\section{Preliminaries}\label{sec:preliminaries}

\paragraph{Problem setup}
Given a dataset $\mathcal{D}=\{x_i,y_i\}^{N}_{i=1}$ where $y_i$ belongs to label space $\mathcal{Y}=\{1,\cdots,K\}$, the objective of machine learning can be formularized as:
\begin{equation}
\theta=\arg\min_{\theta}\sum_{(x_i,y_i) \in \mathcal{D}} \mathcal{L}(f_{\theta}(x_i),y_i)
\label{eq:machinelearningobjective}
\end{equation}
where $f_{\theta}$ is the DNN model parameterized by $\theta$, and $\mathcal{L}$ denotes the training loss function. 
In the scenario of machine unlearning, the goal is to remove the influence of a designated forget set $\mathcal{D}_f \subset \mathcal{D}$ from the pre-trained model $f_{\theta}$, while preserving its performance on the retain subset $\mathcal{D}_r=\mathcal{D} \setminus \mathcal{D}_f$. This process yields the unlearned model $f_{\theta'}$, where $\theta'$ represents the updated parameters. \par

Building upon prior studies \citep{alzubaidi2021review, allen2023backward}, deep learning models are widely recognized to process and interpret inputs in a hierarchical manner, wherein progressively deeper layers encode increasingly abstract and semantically rich representations. As the network depth increases, model parameters are able to capture more complex feature compositions that correspond to higher-level semantic concepts.
Motivated by this perspective, we hypothesize that during pre-training, the model learns a series of feature patterns and their corresponding mappings to semantic concepts. This process can be formalized as $\mathcal{P}_k \mapsto \mathcal{C}_k, k\in[1,K] $, where $\mathcal{P}_k=\{p_{k_1},\cdots,p_{k_M}\}$ denotes the features of the semantic concept $\mathcal{C}_k$ from class $k$.\par

As shown in Figure \ref{fig:FeatureAndConcept}, different concepts may share similar feature patterns, which add to the complexity of the unlearning process. Thereby, we further straighten out such connections among concepts. For two distinct concepts $\mathcal{C}_i$ and $\mathcal{C}_j$, we define the intersection of their feature patterns $\mathcal{P}^{ass}_{ij}=\mathcal{P}_i \cap \mathcal{P}_j$ as the \textbf{associated features}, while the \textbf{unique features} of concept $\mathcal{C}_i$ and $\mathcal{C}_j$ are defined as $\mathcal{P}^{uni}_i=\mathcal{P}_i \setminus \mathcal{P}^{ass}_{ij}$ and $\mathcal{P}^{uni}_j=\mathcal{P}_j \setminus \mathcal{P}^{ass}_{ij}$ other than the shared part.

\paragraph{Unlearning via training with error-maximizing noise}
The previous method, UNSIR, endeavors to achieve unlearning through deliberately inducing catastrophic forgetting about specific target data. This process is achieved through a two-step paradigm:
\begin{equation}
\begin{array}{l}
\theta'_{impair} = \arg\min_{\theta'} \sum_{(x_i,y_i) \in \mathcal{D}_r} \mathcal{L}(f_{\theta}(x_i+\mathcal{N}_f),y_i) \\
\theta'_{repair} = \arg\min_{\theta'} \sum_{(x_i,y_i) \in \mathcal{D}_r} \mathcal{L}(f_{\theta}(x_i),y_i)
\end{array}
\label{eq:UNSIR unlearning steps}
\end{equation}
where $\mathcal{N}_f$ denotes Error-maximizing noise. It is a matrix of the same size as that of the model input, randomly initialized and then trained from the frozen pretrained model through the process:
\begin{equation}
    \mathcal{N}_f = \arg\min_{\mathcal{N}}\mathbb{E} 
    (-\mathcal{L}(f_{\theta}(\mathcal{N}),y^p_i)+\lambda \| w_{noise} \| )
\label{eq:UNSIR error-maximizing noise}
\end{equation}

In this manner, $\mathcal{N}_f$ extracts counteracting feature information that is semantically opposite to the target data from the frozen pretrained model. Under the classification loss $\mathcal{L}$, these anti-feature representations can induce gradient update directions that oppose to those driven by the target data. 
Consequently, during the impair stage in Equation \ref{eq:UNSIR unlearning steps}, when $\mathcal{N}_f$ is
incorporated together with the retain data $\mathcal{D}_r$ and optimized with respect to $y_i$, the model’s originally well-aligned optimization trajectory for target-related features is deliberately disrupted. As a result, the model’s ability to accurately capture and exploit features associated with the target data is significantly degraded, thereby effectively realizing concept-level forgetting of the target data.\par


However, this approach does not account for feature entanglement across different semantic concepts. The error-maximizing noise indiscriminately extracts all anti-feature representations associated with the target data. Directly using these representations for model fine-tuning without proper selection or disentanglement risks disrupting the model’s understanding of features relevant to the retained data. Moreover, this strategy operates solely at the level of feature representations. Owing to the absence of explicit manipulation of the target data’s semantic concepts during the unlearning stage, the mapping from feature patterns to the target semantic concept may still persist in the model’s latent space, ultimately resulting in incomplete and insufficient forgetting.

\section{Methodology}\label{sec:methodology}

In this section, we introduce MeGU for effective machine unlearning, which combines two components: 1) the MLLM guided label perturbation and 2) the Fragment-Align strategy.
As shown in Figure \ref{fig:framework}, we first estimate inter-concept similarities using zero-shot MLLMs on a subset of training data, and cache them in a lightweight transition matrix. Further, based on the pretrained model’s predictions, the transition matrix can assign proper label perturbation to achieve manipulation of concepts. For each forgetting instance, the pre-trained and cached feature noises are injected to disentangle their influence. Furthermore, we utilize MLLM guidance to quantify inter-concept distances to ensure effective label perturbation and identification of the class for positive noise, thereby preserving the retained data.
In the following sections, we first demonstrate the process to generate perturbing labels in Section \ref{subsec:perturbing label generation}. Then, in Section \ref{subsec:feature disentanglement and fine-tuning}, we carefully elucidate the Fragment-Align strategy for influence disentanglement and the fine-tuning method to unlearn target data.

\subsection{Perturbing labels generation}\label{subsec:perturbing label generation}

Intuitively, to induce model unlearning, MeGU introduces semantically consistent but incorrect labels at the finetuning stage to replace the learned connections of target data to their correct labels with a meaningful but alternative relation to the retained concepts. To achieve this, we leverage MLLM to estimate the semantic consistency among concepts to assign suitable perturbing labels.
To reduce the computation costs, we use a light-weight transition matrix to capture inter-concept similarities derived by the MLLM using a small subset from the training data. This matrix, encoding the knowledge of the MLLM, can be repetitively used for perturbing label assignment without further MLLM calls. 

\paragraph{\textit{Transition matrix estimation}}
Specifically, let $q_w$ be the MLLM model parameterized by $w$. We randomly select $n$ exemplars from each class in the dataset $\mathcal{D}$ to construct the subset $\mathcal{D}_{ex}$. Then we prompt the MLLM model $q_w$ to estimate feature similarity of each instance in $\mathcal{D}_{ex}$ to all other semantic concepts. For example, given a query image from class $k$ and its counterpart $l \in [1,K]$, the prompt is:

\vspace{0mm}
\begin{tcolorbox}
\small
Question: This image $<$IMG$_{l}$$>$ shows a photo of $<$label$_{l}>$, True or False? Answer: True;\\
\vspace{-2.5mm}

Question: This image $<$IMG$_{p}$$>$ shows a photo of $<$label$_{l}>$, True or False? Answer: False;\\
\vspace{-2.5mm}

Question: This image $<$IMG$^{query}_{k}$$>$ shows a photo of $<$label$_{l}>$, True or False? Answer:
\end{tcolorbox}
\vspace{-0mm}

In this way, the MLLM output is restricted to binary answers, i.e., True or False. Then, the feature similarity can be represented by softmax output logits.

To compute the feature similarity matrix among concepts, let $x_i$ be an instance from class $k$ in $\mathcal{D}_{ex}$, and $q_w(\cdot)$ be the MLLM output confidence. The feature similarity of $x_i$ with another concept $l$ can be represented as:
\begin{equation}
s_{kl}=q_w(x_i,l)
\label{eq:instance similarity}
\end{equation}    
Then the feature similarity between two concepts labeled $k$ and $l$ can be approximated as:
\begin{equation}
\mathcal{S}_{kl}=\mathbb{E}(q_w(x_i,l)), (x_i,k) \in \mathcal{D}_{ex}
\label{eq:concept similarity}
\end{equation}

Subsequently, the transition possibilities of class $k$ with all other concepts can be calculated through: $\mathbf{t}_k=(\mathcal{S}_{k1}, \mathcal{S}_{k2}, \cdots, \mathcal{S}_{kK})^T/\sum^K_i\mathcal{S}_{ki}$. And the transition matrix $\mathbf{T}$, encoding all inter-conceptional similarities in $\mathcal{D}$, can be obtained via concatenation:
\begin{equation}
\mathbf{T}=(\mathbf{t}_1, \mathbf{t}_2, \cdots, \mathbf{t}_K)
\label{eq:transition matrix}
\end{equation}

\begin{wrapfigure}{r}{0.4\textwidth}
    \centering
    \vspace{-3mm}
    \includegraphics[width=1\linewidth]{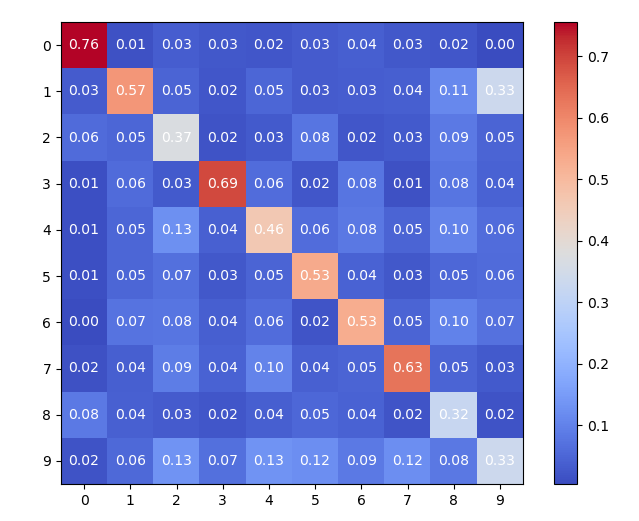}
    \vspace{-5mm}
    \caption{The visualized transition matrix.}
    \vspace{3mm}
    \label{fig:transition matrix}
\end{wrapfigure}

Figure \ref{fig:transition matrix} illustrates the transition matrix of CIFAR-10 as an example.
Although it reflects the proportion of overlapped features among different concepts, this is not adequate for label re-assignment as a given instance does not necessarily express all the features belonging to its concept. Simply applying $\mathbf{T}$ to assign perturbing labels will lead to biases followed by overfitting during unlearning. Therefore, it is essential to take the individual conditions of each instance into account.

\paragraph{\textit{Assigning perturbing labels}}
Given the transition matrix $\mathbf{T}$, the overlapping of an instance $x_i$ from class $k$ with other concepts at the feature level can be computed via:
\begin{equation}
\mathbf{R}_i=\mathbf{T} \cdot (\Tilde{\mathbf{I}}_k \cdot f_{\theta}(x_i))
\label{eq:ranking similarity}
\end{equation}
where $\Tilde{\mathbf{I}}_k$ denotes an identity matrix with the elements of $k$-th row elements set to zero to avoid the influence of the original concept. And $f_{\theta}(x_i)$ denotes the softmax output logits of the pretrained model. \par

Consequently, the perturbing label for an instance $x_i \in \mathcal{D}_f$ is obtained by ranking elements of $\mathbf{R}_i$:
\begin{equation}
y^p_i=\phi(\mathbf{R}_i,\tau) 
\label{eq:perturbing label}
\end{equation}
where $\phi(,\tau)$ denotes the process of selecting the top $\tau$ similar concept as the perturbing label. Here we set the parameter $\tau$ to control the distance between the target data and its re-assigned concept. As a perturbing concept that is too similar to the original one could fail to segregate the target data from the original embedding distribution, leading to insufficient unlearning. While the hindering of alignment with a too distant perturbing label will possibly cause model overfitting or representation collapse. In our experiments, we set $\tau=0.3$ to ensure such a meaningful and alternative connection between target data and retained concepts. In Section \ref{sec:sensitivity studies}, we conduct detailed sensitivity studies for optimal hyperparameters. \par

To better understand the workflow of perturbing label generation, Algorithm \ref{alg:algorithm} provides the detailed pseudo-code.

\begin{algorithm}[H]
\small
\caption{Transition matrix and perturbing labels.}\label{alg:algorithm}
\begin{algorithmic}[1]
    \Require Instance $x_i$ of class $k$ from subset $\mathcal{D}_{ex}$ selected from training set $\mathcal{D}$, concepts $\mathcal{Y}=\{0, 1, \cdots, K-1\}$, MLLM $q_{w}$.
    \State Let $q_{w}(x,y)$ represent the prompted MLLM output of image $x$ and concept $y$.
    \For {$k \in \mathcal{Y}$}
        \For {$l \in \mathcal{Y}$}
            \State $\tilde{\mathcal{S}}_{kl}=\frac{1}{n} \sum^{n}_{i}[q_w(x_i,l)], (x_i,k) \in \mathcal{D}_{ex}$
            \Comment{Eq. \ref{eq:concept similarity}}
        \EndFor
        \State $\mathbf{t}_k=(\tilde{\mathcal{S}}_{k0}, \tilde{\mathcal{S}}_{k1}, \cdots, \tilde{\mathcal{S}}_{kK-1})^T/\sum^{K-1}_i\tilde{\mathcal{S}}_{ki}$
    \EndFor
    \State $\mathbf{T}=(\mathbf{t}_0, \mathbf{t}_1, \cdots, \mathbf{t}_{K-1})$
    \Require Instance $(x_i,y_i)$ from forget set $\mathcal{D}_{f}$, transition matrix $\mathbf{T}$, class concepts $\mathcal{Y}=\{0, 1, \cdots, K-1\}$, pretrained model $p_{\theta}$, identity matrix $\mathbf{I}$ of size $[K,K]$, constant $\tau$.
    \State Let $\phi(,\tau)$ denotes the process of selecting the index of the $\lfloor K\tau \rfloor$-th largest element.
    \For {$(x_i,y_i) \in \mathcal{D}_{f}$}
        \State $\tilde{\mathbf{I}}_{y_i} \gets \mathbf{I}[j,y_i]=0, j \in [0,K-1]$
        \State $\mathbf{R}_i=\mathbf{T} \cdot [\tilde{\mathbf{I}}_{y_i} \cdot p_{\theta}(x_i)]$
        \Comment{Eq. \ref{eq:ranking similarity}}
        \State $y^p_i=\phi(\mathbf{R}_i,\tau) $
    \EndFor
\end{algorithmic}
\end{algorithm}

\par
\paragraph{\textit{The necessity of MLLM guidance for perturbing labels}}
Here, we demonstrate why not use model predictions for label assignment instead of MLLM guidance. Due to the overconfidence, the predicted probabilities of the pretrained model on a class of target data will be concentrated on a very few labels, neglecting the semantic content of individual instances. This poses a risk of leading the pretrained model to overfit to a simple output bias. Conversely, our mechanism, considering individual conditions of target data, guarantees balanced perturbing labels. This facilitates the model in adjusting its decision boundaries according to different instances.
\par


\subsection{Target feature disentanglement and finetuing}\label{subsec:feature disentanglement and fine-tuning}

In Section \ref{sec:introduction}, we demonstrate that the key challenge of unlearning lies in feature entanglement: target data often share features with retained classes, which contributes to over-forgetting. 
To address this problem, we propose the Fragment-Align strategy, which achieves semantic disentanglement through a pair of complementary feature noises. 
During fine-tuning, the \textbf{positive feature noise} $\mathcal{N}^{pos}$ embeds features aligned with the perturbing label, encouraging the target instance to re-anchor toward the new concept and mitigating distributional gaps. 
Conversely, the \textbf{negative feature noise} $\mathcal{N}^{neg}$ suppresses features tied to the original class, facilitating the destruction of target feature patterns in the model's latent space. 
Together, these two noises disentangle the target data from its original concept and realign it with the perturbed concept, facilitating the formation of new decision boundaries and enabling effective forgetting while preserving generalization on retained data.


\paragraph{\textit{Feature noise generation}} Specifically, given an instance from target data, its ground-truth label $\mathcal{C}_{tar}$ and re-assigned perturbing label $\mathcal{C}_{per}$ correspond respectively to feature patterns $\mathcal{P}_{tar}$ and $\mathcal{P}_{per}$. We aim to eliminate the influence of the unique feature $\mathcal{P}^{uni}_{tar}=\mathcal{P}_{tar} \setminus \mathcal{P}^{ass}_{tar|per}$ while preserving shared features $\mathcal{P}^{ass}_{tar|per}=\mathcal{P}_{tar} \cap \mathcal{P}_{per}$ from unexpected disruption. \par
Here, we demonstrate the derivation of feature noises. Let $f_{\theta}$ be the pre-trained model, $(x_i, y_i)$ be an instance from forget data, $y^p_i$ be the perturbing label. The $\mathcal{N}$ is a randomly initialized matrix. The positive feature noise can be obtained through minimizing the loss function towards the perturbing label $y^p_i$ while freezing the pretrained model:

\begin{equation}
    \mathcal{N}^{pos}_i = \arg\min_{\mathcal{N}}\mathbb{E} 
    (\mathcal{L}(f_{\theta}(\mathcal{N}),y^p_i)+\lambda \| w_{noise} \| )
    \label{eq:positive feature noise}
\end{equation}    

Similarly, the negative feature noise can be obtained by changing the training direction towards the original label $y_i$ with a conversed loss function:
\begin{equation}
    \mathcal{N}^{neg}_i = \arg\min_{\mathcal{N}}\mathbb{E}
    (-\mathcal{L}(f_{\theta}(\mathcal{N}),y_i)+\lambda \| w_{noise}\| )
    \label{eq:negative feature noise}
\end{equation}  

Through this process, $\mathcal{N}^{pos}_i$ gradually aligns with the direction of gradient descent, thereby capturing and perturbing features associated with the noisy label $y_i^{p}$. On the other hand, $\mathcal{N}^{neg}_i$ is driven toward the direction of gradient ascent in the model’s optimization landscape to extract counteracting feature information opposite to the target data from the pretrained model. 
When jointly used with the target forgetting data during model re-optimization, the perturbing-label–related features embedded in $\mathcal{N}^{pos}_i$ strengthen the semantic association between the target data features and the perturbed label concept, helping construct alternative connections other than those originally learned during model pre-training. 
Meanwhile, as the model's behavior to $\mathcal{N}^{neg}_i$ is re-optimized in a gradient descent manner, its capability to recognize features associated with the target concept is therefore disrupted.\par

\paragraph{\textit{Model re-optimization}} Intuitively, both the unique and associated feature patterns from concept $y_i$ are restrained by $\mathcal{N}^{neg}_i$. Simultaneously, the associated feature patterns shared with $y^p_i$ are restored by $\mathcal{N}^{pos}_i$. Therefore, by adjusting the relative weights of $\mathcal{N}^{pos}_i$ and $\mathcal{N}^{neg}_i$, the injected feature noises can be precisely controlled to suppress the unique features of the target data, preserve the associated features shared between the target data and the perturbed concept, and introduce new perturbing-label-related features that facilitate aligning the model’s input with the perturbed labels.

Specifically, incorporated with perturbing labels, the perturbed forget dataset is formulated as $\mathcal{D}^p_f=\{(x_i+\mathcal{N}_i, y^p_i)|x_i \in \mathcal{D}_f \}$, where $\mathcal{N}_i$ represents the weighted combination of positive and negative noises:
\begin{equation}
\mathcal{N}_i=\alpha \mathcal{N}^{pos}_i + (1-\alpha) \mathcal{N}^{neg}_i
\label{eq:weighted sum of noises}
\end{equation}
where the constant $\alpha$ serves as a control parameter determining the weights of feature noises. 
Empirically, we set $\alpha=0.7$ according to the results of sensitivity study in Section \ref{sec:sensitivity studies}. 
Note that the feature noises can be pre-generated prior to the unlearning stages and reused across subsequent iterations to reduce computational and time costs. \par

At the finetuning stage, we aim to change the model optimization direction shown in Equation \ref{eq:machinelearningobjective} through a perturbed dataset $\mathcal{D}^p_f$, where representations of target data are pulled away from the original distribution and towards the retained data points.  
Specifically, the unlearning objective is achieved through updating model parameters where $\theta'$ denotes the updated parameters:
\begin{equation}
\begin{aligned}
\theta' = \arg\min_{\theta'} \left\{
\begin{array}{l}
\sum_{(x_i,y^p_i) \in \mathcal{D}^p_f} \mathcal{L}(f_{\theta}(x_i+\mathcal{N}_i),y^p_i) \\
\sum_{(x_i,y_i) \in \mathcal{D}_r} \mathcal{L}(f_{\theta}(x_i),y_i)
\end{array}
\right.
\end{aligned}
\label{eq:MeGU unlearning objective}
\end{equation}

This process reshapes the decision boundaries with alternative embedding distributions. At the same time, the combination of positive and negative feature noises prevents the realignment from disrupting other data points via the disentanglement of target data features, where the positive feature noise additionally smooths the re-optimization.
In Section \ref{sec:ablation studies}, we further explore the effects of these two components and the distinct roles of the two kinds of feature noises respectively through ablation studies.\par
\section{Experiments}\label{sec:experiments}

\begin{table}[t]
\setlength{\tabcolsep}{10pt}
\caption{Class-wise unlearning on CIFAR-100.}\label{table:class-wise CIFAR-100}
\centering
\begin{tabular}{l|cccccccccc}
\toprule[1.3pt]
model              & \multicolumn{1}{c|}{class}                                        & metric & baseline & retrain & FT      & UNSIR & AMNC      & SSD            & \textbf{MaGA}     \\ \hline
\multirow{6}{*}{RN18} & \multicolumn{1}{c|}{\multirow{3}{*}{RKT}} & $A_r$   & 76.30    & 76.19   & 65.43         & 73.83 & 73.59         & \textbf{75.86} & 75.75          \\
                      & \multicolumn{1}{c|}{}                        & $A_f$   & 82.81    & 0.00    & \textbf{0.00} & 41.15 & \textbf{0.00} & \textbf{0.00}  & \textbf{0.00}  \\
                      & \multicolumn{1}{c|}{}                        & MIA    & 96.61    & 8.06    & 10.04         & 3.08  & 28.62         & 0.66           & \textbf{0.00}  \\ \cline{2-10} 
                      & \multicolumn{1}{c|}{\multirow{3}{*}{MR}}     & $A_r$   & 76.38    & 76.23   & 63.90         & 74.26 & 73.22         & 76.20          & \textbf{76.25} \\
                      & \multicolumn{1}{c|}{}                        & $A_f$   & 82.03    & 0.00    & \textbf{0.00} & 8.07  & \textbf{0.00} & \textbf{0.00}  & \textbf{0.00}  \\
                      & \multicolumn{1}{c|}{}                        & MIA    & 95.65    & 6.01    & 12.22         & 1.68  & 46.06         & 0.25           & \textbf{0.00}  \\ \bottomrule[1.3pt]
\multirow{6}{*}{ViT}  & \multicolumn{1}{c|}{\multirow{3}{*}{RKT}} & $A_r$   & 92.27    & 92.04   & 84.26         & 90.83 & 90.53         & 91.39          & \textbf{92.34} \\
                      & \multicolumn{1}{c|}{}                        & $A_f$   & 93.14    & 0.00    & \textbf{0.00} & 24.57 & \textbf{0.00} & \textbf{0.00}  & \textbf{0.00}  \\
                      & \multicolumn{1}{c|}{}                        & MIA    & 84.88    & 6.29    & 16.00         & 11.43 & 1.06          & 6.62          & \textbf{0.00}  \\ \cline{2-10} 
                      & \multicolumn{1}{c|}{\multirow{3}{*}{MR}}     & $A_r$   & 92.20    & 92.18   & 84.15         & 89.95 & 89.95         & 91.78         & \textbf{92.33}         \\
                      & \multicolumn{1}{c|}{}                        & $A_f$   & 98.44    & 0.00    & \textbf{0.00} & 56.25 & \textbf{0.00} & \textbf{0.00}  & \textbf{0.00}  \\
                      & \multicolumn{1}{c|}{}                        & MIA    & 90.24    & 0.86    & 5.05          & 2.61  & 0.88          & 1.45          & \textbf{0.00}  \\
                      \bottomrule[1.3pt]
\end{tabular}
\vspace{-4mm}
\end{table}

\begin{table}[t]
\setlength{\tabcolsep}{10pt}
\vspace{3mm}
\caption{Class-wise unlearning on CIFAR-20.}\label{table:class-wise CIFAR-20}
\centering
\begin{tabular}{l|cccccccccc}
\toprule[1.3pt]
model              & \multicolumn{1}{c|}{class}                                        & metric & baseline & retrain & FT      & UNSIR & AMNC      & SSD            & \textbf{MaGA}     \\ \hline
\multirow{6}{*}{RN18} & \multicolumn{1}{c|}{\multirow{3}{*}{veh2}} & $A_r$   & 82.84    & 82.41   & 73.74         & 81.41 & \textbf{82.21}         & 83.38 & 82.93          \\
                      & \multicolumn{1}{c|}{}                        & $A_f$   & 84.99    & 0.00    & \textbf{0.00} & 58.82 & \textbf{0.00} & 22.37          & \textbf{0.00}  \\
                      & \multicolumn{1}{c|}{}                        & MIA    & 87.72    & 14.24   & 40.84         & 45.68 & 7.84          & 5.72           & \textbf{0.08}  \\ \cline{2-10} 
                      & \multicolumn{1}{c|}{\multirow{3}{*}{veg}}     & $A_r$   & 82.59    & 82.24   & 72.56         & 81.23 & 81.66         & 82.64         & \textbf{82.53} \\
                      & \multicolumn{1}{c|}{}                        & $A_f$   & 88.94    & 0.00    & \textbf{0.00} & 70.24 & \textbf{0.00} & 45.34          & \textbf{0.00}  \\
                      & \multicolumn{1}{c|}{}                        & MIA    & 93.28    & 9.24    & 29.16         & 43.27  & 3.04          & 2.08           & \textbf{0.00}  \\ \bottomrule[1.3pt]
\multirow{6}{*}{ViT}  & \multicolumn{1}{c|}{\multirow{3}{*}{veh2}} & $A_r$   & 96.08    & 95.35   & 85.13         & 93.83 & 94.20         & 90.26          & \textbf{95.82} \\
                      & \multicolumn{1}{c|}{}                        & $A_f$   & 94.35    & 0.00    & \textbf{0.00} & 67.29 & \textbf{0.00} & \textbf{0.00}  & \textbf{0.00}  \\
                      & \multicolumn{1}{c|}{}                        & MIA    & 80.96    & 20.36   & 22.00          & 49.96 & 1.26           & 9.88            & \textbf{0.00}  \\ \cline{2-10} 
                      & \multicolumn{1}{c|}{\multirow{3}{*}{veg}}     & $A_r$   & 95.88    & 94.95   & 88.52         & 93.75 & 93.31         & \textbf{95.61}          & 95.64 \\
                      & \multicolumn{1}{c|}{}                        & $A_f$   & 97.67    & 0.00    & 0.59 & 86.57 & \textbf{0.00} & \textbf{0.00}  & \textbf{0.00}  \\
                      & \multicolumn{1}{c|}{}                        & MIA    & 91.48    & 4.16    & 14.44          & 63.6  & 1.05           & 1.44           & \textbf{0.00}  \\ \bottomrule[1.3pt]
\end{tabular}
\vspace{0mm}
\end{table}

In this section, we demonstrate experimental results on different unlearning tasks and datasets and compare the performance of MaGA with existing methods using a set of metrics. We randomly sample 10,000 instances from the retain dataset $\mathcal{D}_r$ along with the perturbed forget set $\mathcal{D}^{p}_{f}$ for fine-tuning using Eq. \ref{eq:MeGU unlearning objective}. We perform 3 epochs of unlearning for ResNet18, and 5 epochs for Vision Transformer (ViT). While increasing the number of unlearning epochs will further enhance performance, it comes at the expense of computational efficiency. Note that the metrics in tables are represented as percentages, and the boldings indicate superiority. For detailed hyper-parameter settings, we conduct sensitivity in Appendix \ref{sec:sensitivity studies}. To better understand the effects of perturbing labels and feature noises, we conduct ablation studies in the Appendix \ref{sec:ablation studies} by comparing MaGA with a randomly-selecting perturbing label strategy and a none feature noise framework. We also visualize the predictions of unlearned models compared with the "gold model" and baseline model to demonstrate the effects of unlearning in Appendix \ref{sec:visualization output closeness}. More results of class-wise and sub-class unlearning experiments on other target data are also present in Appendix \ref{sec:supplementary experiments}. 

\subsection{Experimental setup}\label{sec:experimental setup}

\textbf{Datasets:} Following \citep{foster2024fastSSD}, we evaluate our proposed method for image classification models using CIFAR-10, CIFAR-100 \citep{krizhevsky2009learningCIFAR10}, CIFAR-20 \citep{xie2020unsupervisedCIFAR20}.
\par
\textbf{Models:} We leverage MMICL \citep{mmicl_zhao2023} as our MLLM zero-shot machine expert. We conduct unlearning experiments on two types of backbones: ResNet18 \citep{he2016deep} and Vision Transformer (ViT) \citep{dosovitskiy2020image}. Models are trained using Adam optimizer \citep{diederik2014adam} and a multi-step scheduler with the initial learning rate set to 0.1. In convenience of comparisons, we control the training of the baseline model to align with that of the previous studies. The pretraining and unlearning processes are carried out on NVIDIA RTX4090 and Intel Xeon platform. The memory usage is controlled within 30 GiB for MLLM inference and less for unlearning.
\par
\textbf{Unlearning tasks:} Following previous studies \citep{foster2024fastSSD, chundawat2023canbadteacher, tarun2023fastUNSIR}, we evaluate the effectiveness of MaGA across three distinct unlearning tasks, including: 1) Class-wise unlearning conducted on CIFAR-100 and CIFAR-20, which unlearns an entire superclass from the dataset. 2) Sub-class unlearning conducted on CIFAR-20, where a sub-class within a superclass is forgotten. 3) Random unlearning conducted on CIFAR-10 where a subset randomly selected from the original dataset is forgotten.
\par
\textbf{Baselines:} Following previous researches \citep{foster2024fastSSD, graves2021amnesiac}, we compare our method with baselines including: "\textit{baseline}": the original pre-trained model, "\textit{retrain}": the gold model trained from scratch, "\textit{FT}" which finetunes the pre-trained model for 5 epochs with only retain data, "\textit{UNSIR}" \citep{tarun2023fastUNSIR}, "\textit{teacher}" which denotes Bad Teacher \citep{chundawat2023canbadteacher}, "\textit{AMNC}" which denotes Amnesiac \citep{graves2021amnesiac}, and "\textit{SSD}" \citep{foster2024fastSSD}.
\par
\paragraph{\textbf{Evaluation metrics:}} Following \citep{chundawat2023canbadteacher}, we use: 1) Accuracy on forget and retain set, denoted as $A_f$ and $A_r$ respectively; 2) Membership Inference Attack (MIA) score \citep{shokri2017membership} as evaluating metrics. 
Considering the Streisand effect \citep{chundawat2023canbadteacher}, we demonstrate that the unlearned model that aligns close to the gold model on accuracy is "well-unlearned". Meanwhile, lower MIA probabilities indicate better security in adversary attacks. \par


\subsection{Main implementation results}\label{sec:main implementation results}

\begin{table}[t]
\setlength{\tabcolsep}{10pt}
\caption{Sub-class unlearning on CIFAR-20.}\label{table:sub-class CIFAR-20}
\centering
\begin{tabular}{l|cccccccccc}
\toprule[1.3pt]
model              & \multicolumn{1}{c|}{sublass}                                        & metric & baseline & retrain & FT & UNSIR & AMNC       & SSD   & \textbf{MaGA}     \\ \hline
\multirow{6}{*}{RN18} & \multicolumn{1}{c|}{\multirow{3}{*}{RKT}} & $A_r$   & 82.93    & 83.55   & 73.57    & 81.50 & 81.89          & 82.36 & \textbf{82.44} \\
                      & \multicolumn{1}{c|}{}                        & $A_f$   & 81.51    & 1.39    & 15.36    & 60.85 & \textbf{0.00}  & 5.90  & 4.51           \\
                      & \multicolumn{1}{c|}{}                        & MIA    & 84.43    & 22.40   & 12.29    & 43.6  & 9.41           & 5.85  & \textbf{4.80}  \\ \cline{2-10} 
                      & \multicolumn{1}{c|}{\multirow{3}{*}{sea}}     & $A_r$   & 82.79    & 82.95   & 85.85    & 81.14 & 82.06          & 82.41 & \textbf{82.80} \\
                      & \multicolumn{1}{c|}{}                        & $A_f$   & 97.66    & 77.34   & \textbf{76.14}    & 95.49 & 35.76 & 86.02 & 84.38          \\
                      & \multicolumn{1}{c|}{}                        & MIA    & 87.10    & 56.67   & 66.05    & 86.60 & \textbf{4.00}  & 54.65 & 66.68          \\ \bottomrule[1.3pt]
\multirow{6}{*}{ViT}  & \multicolumn{1}{c|}{\multirow{3}{*}{RKT}} & $A_r$   & 96.01    & 96.10   & 88.61    & 93.59 & 92.39          & 96.18 & \textbf{96.05} \\
                      & \multicolumn{1}{c|}{}                        & $A_f$   & 94.70    & 2.83    & 5.12     & 66.75 & 0.78           & 22.83 & \textbf{1.10}  \\
                      & \multicolumn{1}{c|}{}                        & MIA    & 85.80    & 7.44    & 19.41    & 15.47 & 0.86           & 3.45  & \textbf{0.00}  \\ \cline{2-10} 
                      & \multicolumn{1}{c|}{\multirow{3}{*}{sea}}     & $A_r$   & 95.97    & 95.55   & 88.47    & 94.28 & 93.82          & 95.96 & \textbf{96.05} \\
                      & \multicolumn{1}{c|}{}                        & $A_f$   & 98.44    & 74.22   & 81.68    & 89.15 & 17.97          & 97.66 & \textbf{12.76} \\
                      & \multicolumn{1}{c|}{}                        & MIA    & 89.26    & 41.43   & 43.27    & 69.20 & 0.22           & 86.49 & \textbf{0.20}  \\ \bottomrule[1.3pt]
\end{tabular}
\vspace{-4mm}
\end{table}

\begin{table}[t]
\vspace{3mm}
\caption{Random unlearning on CIFAR-10.}\label{table:random CIFAR-10}
\centering
\setlength{\tabcolsep}{13pt}
\begin{tabular}{l|cccccccccc}
\toprule[1.3pt]
model                                   & metric & baseline & retrain & FT & teacher & AMNC       & SSD   & \textbf{MaGA}    \\ \hline
\multicolumn{1}{c|}{\multirow{3}{*}{RN18}} & $A_r$   & 95.31    & 92.42   & 87.86    & 90.68   & 90.63          & 91.38 & \textbf{93.04} \\
\multicolumn{1}{c|}{}                      & $A_f$   & 94.03    & 94.40   & 88.77    & \textbf{93.13}   & 55.35 & 95.88 & 85.59          \\
\multicolumn{1}{c|}{}                      & MIA    & 76.53    & 74.42   & 71.70    & 47.08   & \textbf{17.89} & 76.50 & 34.04          \\ \bottomrule[1.3pt]
\multicolumn{1}{c|}{\multirow{3}{*}{ViT}}  & $A_r$   & 98.66    & 98.75   & 97.64    & 98.17   & 98.15          & 98.63 & \textbf{98.81} \\
\multicolumn{1}{c|}{}                      & $A_f$   & 99.71    & 99.41   & 98.24    & 93.52   & 74.88 & \textbf{99.39} & 96.27          \\
\multicolumn{1}{c|}{}                      & MIA    & 88.66    & 91.67   & 86.83    & 26.80   & \textbf{6.40}  & 88.22 & 29.18          \\ \bottomrule[1.3pt]
\end{tabular}
\vspace{0mm}
\end{table}

\paragraph{\textbf{Class-wise unlearning:}} Experiments are conducted on CIFAR-100 and CIFAR-20 using ResNet18 and Vision Transformer as classification backbones. For CIFAR-100, we designate two classes: \textit{rocket}  (denoted as "RKT") and \textit{mushroom} (denoted as "MR"). The results are shown in Table \ref{table:class-wise CIFAR-100}. Metrics demonstrate that our method successfully aligns its performance with the "gold model" denoted as "retrain". Especially when class-wisely unlearning \textit{rocket} using ViT, MaGA narrows the gap with retrained model in retain accuracy by 0.35\% compared to \textit{SSD}.
For CIFAR-20, we forget two superclasses: \textit{Vehicle2} and \textit{veg}. MaGA evidently lowers the retain accuracy compared with previous methods such as \textit{SSD}, particularly decreasing over 40\% on \textit{vegetable} using ResNet18. Meanwhile, MaGA still maintains a competitive retain accuracy of 82.53\%, which is closer to 82.24\% of the "gold model". Thus, it is demonstrated that MaGA achieves a balance between complete unlearning and preservation of overall generalization.
Crucially, MaGA significantly lowers the MIA risk compared to existing methods on most tasks, with 9.8\% and 49\% lower than \textit{SSD} and \textit{UNSIR} respectively on \textit{veh2} from CIFAR-20 using ViT. This highlights the effectiveness of removing target information without being recognized by adversaries. 
\par

\paragraph{\textbf{Sub-class unlearning:}} We perform unlearning on two sub-classes: \textit{rocket} and \textit{sea}, belonging to the super-classes "\textit{Vehicle2}" and "\textit{natural scenes}", respectively. The results are summarized in Table \ref{table:sub-class CIFAR-20}. The challenge of sub-class unlearning exists that the target sub-class shares feature patterns with fellows within the same super-class. Thus, when the features of the target sub-class are unlearned, the generalization capabilities of the entire super-class are compromised. This is evident in methods such as \textit{UNSIR} and \textit{Amnesiac}, which exhibit significant reductions in retain accuracy. 
In contrast, benefiting from the disentanglement of Fragment-Align strategy, the performance of MaGA on retained data can be well preserved and closely aligned with the "gold model". Taking sub-class \textit{rocket} with ViT backbone as an example, MaGA increases the retain accuracy by 2.51\% and 3.71\% compared to \textit{UNSIR} and \textit{Amnesiac} respectively. 
The superiority of MIA security of MaGA unlearning is also recurrently demonstrated among these cases, especially 86\% and 69\% lower than that of \textit{SSD} and \textit{UNSIR} on \textit{sea} with ViT.

\begin{wrapfigure}{r}{0.5\textwidth}
    \centering
    \vspace{0mm}
    \includegraphics[width=0.95\linewidth]{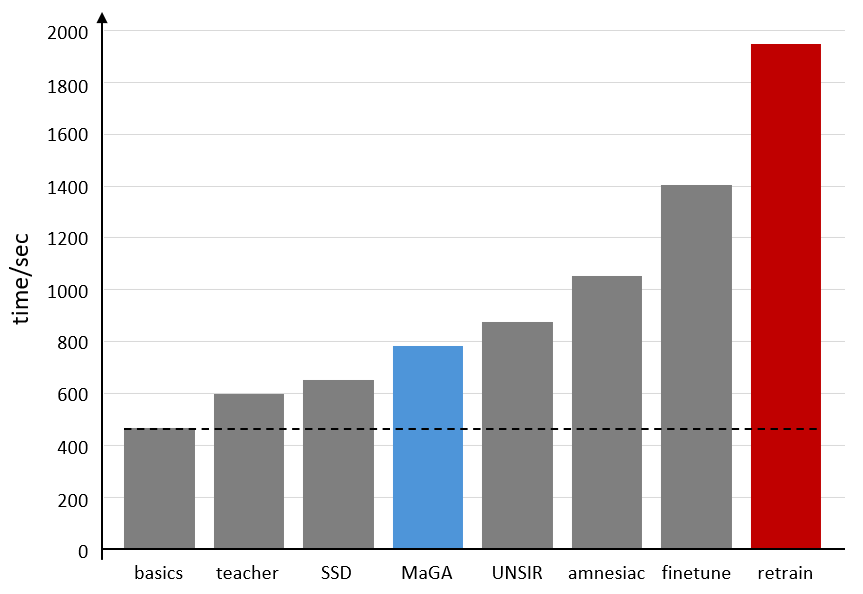}
    \vspace{-2mm}
    \caption{Time consumed for CIFAR-100 class-wise unlearning.}
    \vspace{-2mm}
    \label{fig:time_comparison}
\end{wrapfigure}

\paragraph{\textbf{Random unlearning:}} We utilize the CIFAR-10 dataset for random unlearning, where a set of instances (100 in our experiments) is randomly selected as the target dataset to be forgotten. 
As shown in Table \ref{table:random CIFAR-10}, random unlearning is inherently more difficult as the retrained gold model also persists a relatively high forget accuracy. In random unlearning, the forget set and retain set may contain samples from the same semantic class. Thus the unlearned model preserves generalization capability toward the class, making it possible to correctly classify forgotten samples despite their removal. More importantly, this indicates that, in such settings, the unlearning performance cannot be fully measured by accuracy alone. In this case, MaGA competitively reduces MIA scores compared to most baselines and the gold model, with 40\% and 79\% lower than \textit{SSD} using ResNet18 and ViT respectively, which suggests that the model no longer remembers specific target instances, despite still leveraging general class-level knowledge. Simultaneously, it is still illustrated that MaGA behaves closely to the "gold model", with the overall generalization preserved after unlearning. 

\paragraph{\textbf{Computation time:}} We evaluate the time required to complete the unlearning process for each method, where the "basics" accounts for the time spent on dataset preparation, model loading, and metric computation. Using CIFAR-100 class-wise unlearning with the Vision Transformer (ViT) backbone as an example, MaGA reduces computational time by over 59\% compared to full retraining, as illustrated in Figure \ref{fig:time_comparison}.

\section{Conclusion}\label{sec:conclusion}
\par
\paragraph{\textbf{Contributions:}} In this work, we introduce MeGU, a novel machine unlearning framework that effectively eliminate target data influence while maintaining overall model generalization through pioneering feature disentanglement and data realignment under the guidance of MLLMs. Extensive experiments across diverse datasets and forgetting tasks validate that MeGU consistently outperforms existing unlearning methods under most conditions, ensuring secured and effective unlearning. In future work, this framework can be extended to accommodate more complex unlearning scenarios and a broader range of pre-trained models. 
\par
\paragraph{\textbf{Limitations:}} 
Due to the computation cost of utilizing MLLMs, MeGU is not the most time-efficient unlearning approaches. Nevertheless, this limitation can be mitigated through several strategies: 1) employing zero-shot inference from MLLMs solely as a form of knowledge guidance; 2) precomputing conceptual similarities and the transition matrix for a given dataset prior to the occurrence of any unlearning requests, and subsequently reusing the obtained transition matrix across all future unlearning tasks. Furthermore, the number of exemplars used in estimating the transition matrix can be flexibly adjusted to further reduce computational costs.
\par

\bibliographystyle{unsrtnat}
\bibliography{references}
\appendix
\clearpage

\section*{Appendix}\label{sec:appendix}
\appendix
\renewcommand{\thesubsection}{\Alph{subsection}}

\subsection{Algorithm of proposed MeGU}\label{sec:algorithm}
In Section \ref{sec:methodology}, we propose MeGU as our unlearning method. MeGU employs zero-shot MLLM as machine experts to estimate the feature similarities between different concepts, represented by a transition matrix $\mathbf{T}$. This matrix is subsequently used to assign perturbing labels to the dataset, instructing the utilization of feature noises to disentangle the influence of the target concept. This approach ensures effective unlearning while preserving overall generalization. To better understand the workflow of perturbing label generation, we provide a detailed pseudo-code below.

\begin{algorithm}[H]
\small
\caption{Transition matrix and perturbing labels.}\label{alg:algorithm}
\begin{algorithmic}[1]
    \Require Instance $x_i$ of class $k$ from subset $\mathcal{D}_{ex}$ selected from training set $\mathcal{D}$, concepts $\mathcal{Y}=\{0, 1, \cdots, K-1\}$, MLLM $q_{w}$.
    \State Let $q_{w}(x,y)$ represent the prompted MLLM output of image $x$ and concept $y$.
    \For {$k \in \mathcal{Y}$}
        \For {$l \in \mathcal{Y}$}
            \State $\tilde{\mathcal{S}}_{kl}=\frac{1}{n} \sum^{n}_{i}[q_w(x_i,l)], (x_i,k) \in \mathcal{D}_{ex}$
            \Comment{Eq. \ref{eq:concept similarity}}
        \EndFor
        \State $\mathbf{t}_k=(\tilde{\mathcal{S}}_{k0}, \tilde{\mathcal{S}}_{k1}, \cdots, \tilde{\mathcal{S}}_{kK-1})^T/\sum^{K-1}_i\tilde{\mathcal{S}}_{ki}$
    \EndFor
    \State $\mathbf{T}=(\mathbf{t}_0, \mathbf{t}_1, \cdots, \mathbf{t}_{K-1})$
    \Require Instance $(x_i,y_i)$ from forget set $\mathcal{D}_{f}$, transition matrix $\mathbf{T}$, class concepts $\mathcal{Y}=\{0, 1, \cdots, K-1\}$, pretrained model $p_{\theta}$, identity matrix $\mathbf{I}$ of size $[K,K]$, constant $\tau$.
    \State Let $\phi(,\tau)$ denotes the process of selecting the index of the $\lfloor K\tau \rfloor$-th largest element.
    \For {$(x_i,y_i) \in \mathcal{D}_{f}$}
        \State $\tilde{\mathbf{I}}_{y_i} \gets \mathbf{I}[j,y_i]=0, j \in [0,K-1]$
        \State $\mathbf{R}_i=\mathbf{T} \cdot [\tilde{\mathbf{I}}_{y_i} \cdot p_{\theta}(x_i)]$
        \Comment{Eq. \ref{eq:ranking similarity}}
        \State $y^p_i=\phi(\mathbf{R}_i,\tau) $
    \EndFor
\end{algorithmic}
\end{algorithm}

\subsection{Parameter settings for experiments}\label{sec:parameter settings}
During the unlearning process, different configurations are employed for the ResNet18 and Vision Transformer (ViT) backbones to reach a balance between effectiveness and efficiency. These settings are detailed in Table \ref{table:hyperparameters}. The "noise size" refers to the batch size of feature noise samples extracted per class, which are randomly selected to be integrated with the target data. We at one time train a batch of feature noise for one class to enrich the variety. However, empirical results show that this will not affect the overall effectiveness of our method. Thus, the noise batch size can be decreased to save computation and time costs.

\begin{table}[h]
\vspace{0mm}
\caption{Parameter setting for unlearning.}\label{table:hyperparameters}
\centering
\begin{tabular}{l|cccccccccc}
\toprule[1.3pt]
Parameters                       & ResNet18 & ViT     \\ \hline
exemplar $n$/class               & 10       & 10      \\
batch size                       & 32       & 32      \\
noise size                       & 256      & 64      \\
noise lr                         & 0.1      & 0.1     \\
unlearn lr                       & 0.0003   & 0.00005 \\
rank $\tau$                      & 0.3      & 0.3     \\
proportion $\alpha$              & 0.7      & 0.7     \\
number of instances for finetune & 10000    & 10000   \\
finetune iteration               & 3        & 5       \\
\bottomrule[1.3pt]
\end{tabular}
\vspace{0mm}
\end{table}

\subsection{Distribution of perturbing labels}\label{sec:distribution of perturbing labels}
In Section \ref{subsec:perturbing label generation}, we present a perturbing label assignment strategy grounded in the feature similarities between each instance in the forget set $\mathcal{D}_f$ and the retained concepts. This strategy leverages the conceptual relationships encoded within the transition matrix $\mathbf{T}$, in conjunction with the predictions of $\mathcal{D}_f$ obtained from the unlearned model, to effectively quantify such feature-level affinities.
\par

\begin{wrapfigure}{r}{0.5\textwidth}
    \centering
    \vspace{-4mm}
    \includegraphics[width=1\linewidth]{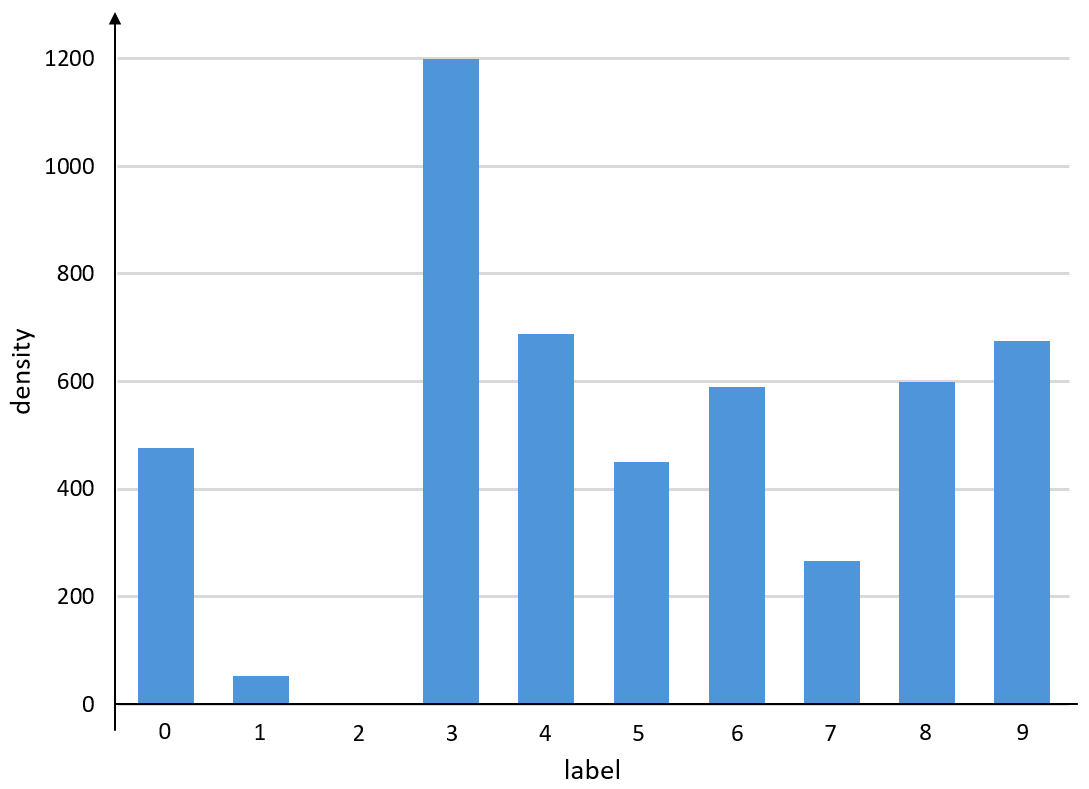}
    \vspace{-8mm}
    \caption{The distribution of perturbing labels for class-wise unlearning designating class 2 as target on CIFAR-10 with ResNet18.}
    \vspace{-1mm}
    \label{fig:label density}
\end{wrapfigure}

In contrast to unbalanced label assignment methods, which often mislead the model by systematically misclassifying target instances into a single erroneous class, our approach introduces diversity in the assignment of perturbing labels across instances. This variation mitigates the risk of inducing model bias and the Streisand effect during fine-tuning, by promoting a balanced and context-aware distribution of perturbing labels.
Moreover, compared with complete random label assignment, our approach systematically considers the semantic compatibility between target instances and their assigned perturbing concepts. This compatibility is quantified through feature similarity, thereby guiding the reassignment process in a principled manner. Such alignment facilitates more effective unlearning, as target instances are more likely to be aligned with semantically related concepts, reducing unintended model disruption.

Figure \ref{fig:label density} visualizes the distribution of perturbing labels assigned to the forget dataset. This experiment is conducted on CIFAR-10, with class 2 designated as the target class, using ResNet18 as the backbone. While the observed disparities in perturbing label frequencies may partially reflect varying conceptual proximities to the target class, the overall distribution remains notably balanced, thereby validating the efficacy of our proposed assignment mechanism.

\subsection{Comparison on visualized model prediction behavior after unlearning}\label{sec:visualization output closeness}
\begin{figure}[h]
    \centering
    \includegraphics[width=0.75\linewidth]{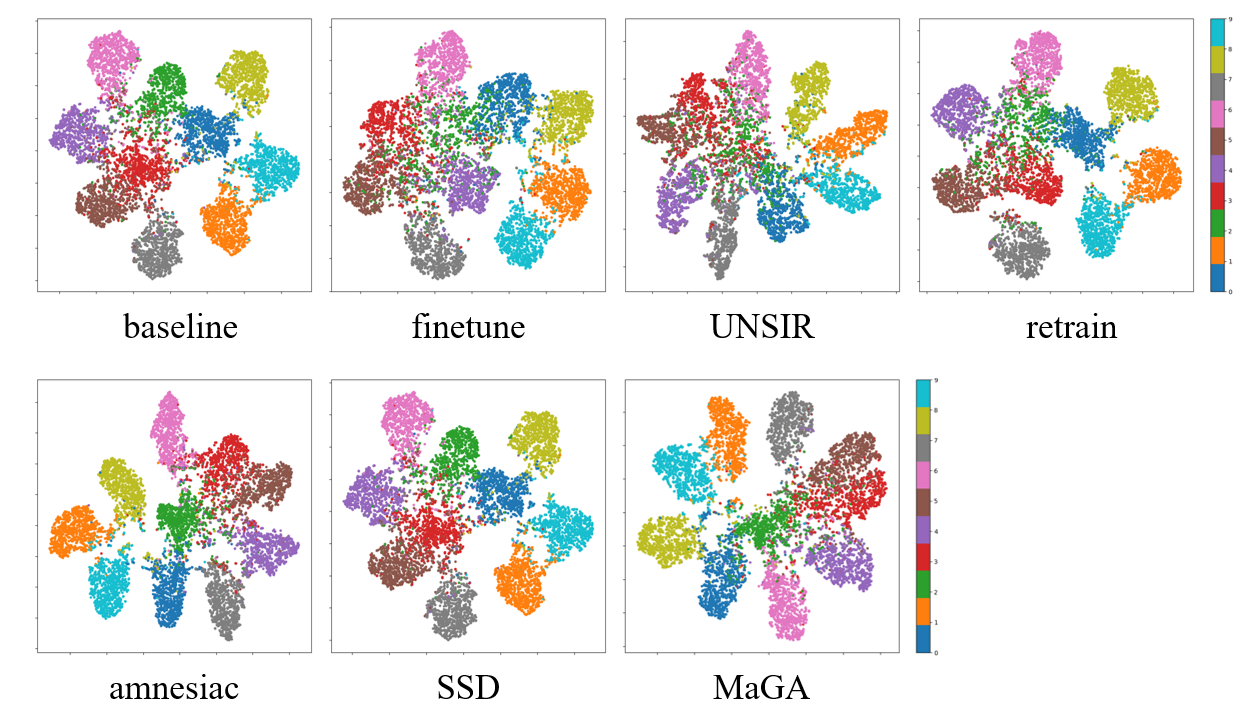}
    \vspace{0mm}
    \caption{Visulized model prediction behavior comparison.}
    \vspace{-2mm}
    \label{fig:model prediction visulization}
\end{figure}

The alignment of the unlearned model with the "gold model" is one of the most significant indicators when evaluating the unlearning performance, as is demonstrated in Section \ref{sec:experimental setup}. In order to perceptually compare our proposed method with baselines, we utilize t-SNE to visualize the predictions of the unlearned model using different methods on CIFAR-10 test set. The results are shown in Figure \ref{fig:model prediction visulization}.

Each point in the figure represents an instance, while each color denotes a label (concept). Theoretically, for a model with better classification performance, the distributions of different concepts are better separated while that of the same concept should be more compact~\cite{hong2025data}. However, in class-wise unlearning scenarios, we expect model generalization on the target class to be disrupted while preserving that on retained classes. Thereby, the distribution of unlearned concept, which is designated to be label 2 represented as green, is expected to be dispersed compared with the original pretrained model (denoted as "baseline"). At the same time, the distributions of retain data are expected to maintain to preserve overall generalization. Comparing the prediction distributions of unlearned models, MeGU behaves more similar to the "gold model", with target instances split and captured by other well-separated retained concepts. This demonstrates the superiority of MeGU considering the closeness to the fully-retrained model.

\begin{figure}[t]
    \centering
    \includegraphics[width=0.6\linewidth]{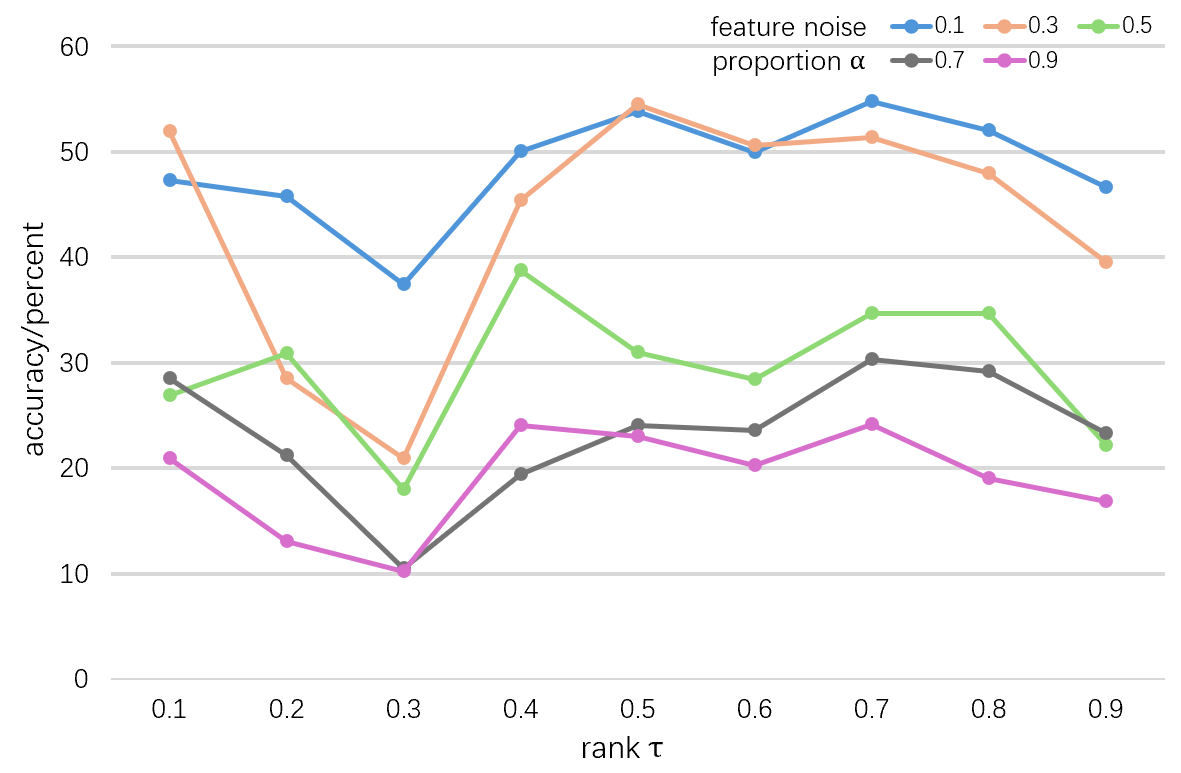}
    \vspace{-2mm}
    \caption{Performance on forget set under varying hyperparameters.}
    \vspace{0mm}
    \label{fig:sensitivity forget}
\end{figure}

\begin{figure}[t]
    \centering
    \includegraphics[width=0.6\linewidth]{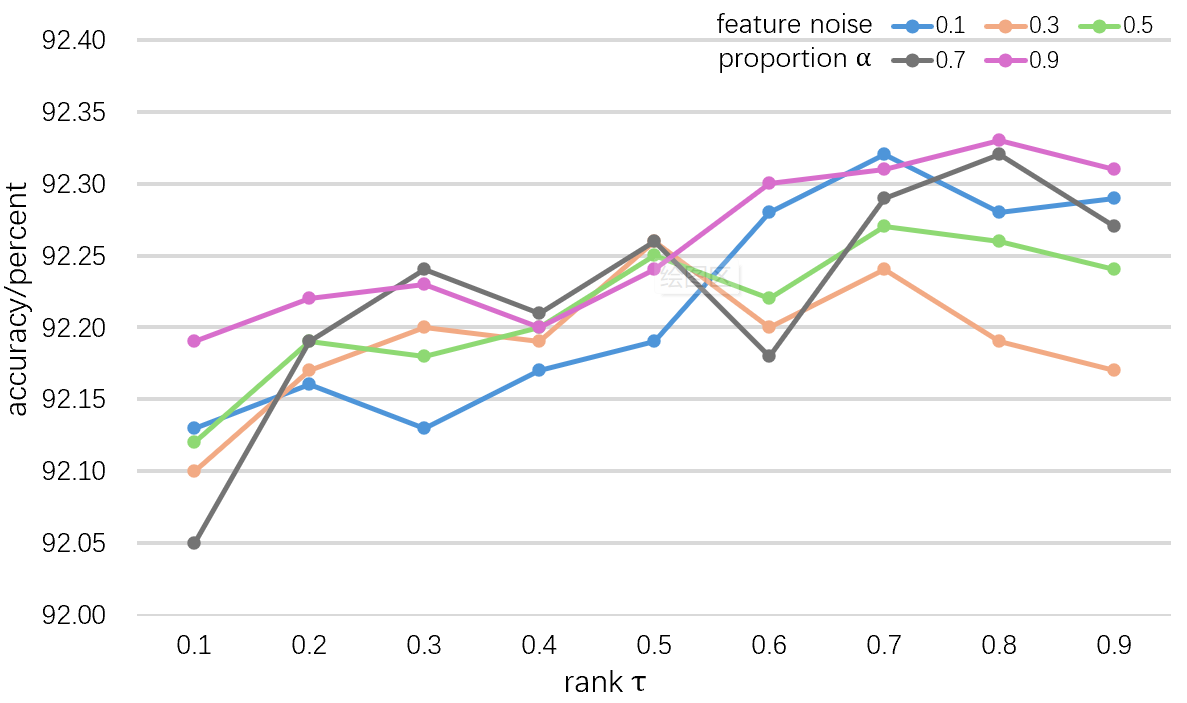}
    \vspace{-2mm}
    \caption{Performance on retain set under varying hyperparameters.}
    \vspace{0mm}
    \label{fig:sensitivity retain}
\end{figure}

\subsection{Sensitivity studies}\label{sec:sensitivity studies}
In our proposed method MeGU, two key hyperparameters, the rank constant $\tau$ and the noise proportion $\alpha$, play pivotal roles in the unlearning process. The rank constant $\tau$ regulates the feature similarity between the assigned perturbing labels and the corresponding target data. This is crucial, as an excessively high similarity between the reassigned and original concepts would result in the preservation of too many associated feature patterns, hindering adequate unlearning.
Meanwhile, it is highlighted that the noise proportion $\alpha$ determines the balance between the positive feature noise encoded from the retaining data and the negative feature noise encoded from the forgetting data. As discussed in Section \ref{subsec:Fragment-Align fine-tuning}, this pair of feature noises poses different impacts on the influence disentanglement of target data during fine-tuning. \par

Here, to fully explore MeGU, we conduct sensitivity studies on these two hyperparameters. We record the class-wise performance of the unlearned model under varying settings using CIFAR-100 dataset. To more explicitly demonstrate the influence of different hyper-parameters on retain accuracy, we manually induce incomplete unlearning by setting the learning rate to 1e-5, which is much lower than our other experiments. We conduct 3 independent experiments for each hyperparameter setting. The averaged accuracy performance is calculated to reflect the general trend. Figure \ref{fig:sensitivity forget} and Figure \ref{fig:sensitivity retain} illustrate the classification accuracy on forget and retain dataset. It is shown that generally, a higher $\alpha$ represents lower accuracy in the retain set, enhancing the effects of positive feature noise on aligning perturbed instances with retain data. At the same time, the trend of retain accuracy under a changing rank $\tau$ is investigated. The prior hypothesis is verified that although we intuitively expect a more similar concept as the perturbing label, such a perturbing label with too many associated feature patterns will hinder the forgetting of target information. On the other hand, there is hardly clear regularity corresponding to $\alpha$ or $\tau$ due to the saturated classification performance on retain data. Thus, in experiments, a $\tau$ around 0.2 to 0.3 and an $\alpha$ around 0.7 to 0.9 are preferred.

\begin{table}[t]
\vspace{0mm}
\caption{Ablation studies on class-wise unlearning with CIFAR-100.}\label{table:ablation}
\vspace{0mm}
\centering
\begin{tabular}{l|ccccc|ccc}
\toprule[1.3pt]
model              & \multicolumn{1}{c|}{class}                                         & Metric & baseline & retrain       & SSD           & RND            & w.o.F.N       & MeGU     \\ \hline
\multirow{6}{*}{RN18} & \multicolumn{1}{c|}{\multirow{3}{*}{rocket}} & $Ar$   & 76.30    & 76.19         & 75.86         & \textbf{76.24} & 75.46         & 75.75          \\
                      & \multicolumn{1}{c|}{}                        & $Af$   & 82.81    & \textbf{0.00} & \textbf{0.00} & 56.51          & 3.32          & \textbf{0.00}  \\
                      & \multicolumn{1}{c|}{}                        & MIA    & 96.61    & 8.06          & 0.66          & 20.4           & \textbf{0.00} & \textbf{0.00}  \\ \cline{2-9} 
                      & \multicolumn{1}{c|}{\multirow{3}{*}{MR}}     & $Ar$   & 76.38    & 76.16         & 76.20         & 75.98          & 76.07         & \textbf{76.25} \\
                      & \multicolumn{1}{c|}{}                        & $Af$   & 82.03    & \textbf{0.00} & \textbf{0.00} & 57.12          & 4.55          & \textbf{0.00}  \\
                      & \multicolumn{1}{c|}{}                        & MIA    & 95.65    & 5.61          & 0.25          & 10.21          & \textbf{0.00} & \textbf{0.00}  \\ \hline
\multirow{6}{*}{ViT}  & \multicolumn{1}{c|}{\multirow{3}{*}{rocket}} & $Ar$   & 92.27    & 91.84         & 91.39         & \textbf{92.37} & 91.96         & 92.34          \\
                      & \multicolumn{1}{c|}{}                        & $Af$   & 93.14    & \textbf{0.00} & \textbf{0.00} & 56.51          & 3.84          & \textbf{0.00}  \\
                      & \multicolumn{1}{c|}{}                        & MIA    & 84.88    & 6.29          & 6.62          & 2.8            & 0.00          & \textbf{0.00}  \\ \cline{2-9} 
                      & \multicolumn{1}{c|}{\multirow{3}{*}{MR}}     & $Ar$   & 92.20    & 92.18         & 91.78         & 92.07          & 91.79         & \textbf{92.33} \\
                      & \multicolumn{1}{c|}{}                        & $Af$   & 98.44    & \textbf{0.00} & \textbf{0.00} & 72.66          & 10.17         & \textbf{0.00}  \\
                      & \multicolumn{1}{c|}{}                        & MIA    & 90.24    & 0.86          & 1.45          & 1.4            & \textbf{0.00} & \textbf{0.00}  \\ \bottomrule[1.3pt]
\end{tabular}
\end{table}

\subsection{Ablation studies}\label{sec:ablation studies}

To deepen the understanding of the intrinsic properties of the MeGU framework and evaluate the impact of its key components: machine-guided perturbing labels and the following pair of feature noises. We conduct ablation studies using various combinations of these elements on class-wise unlearning tasks with CIFAR-100. The results are presented in Table \ref{table:ablation}, where \textit{'RND'} refers to assigning random labels to forget data instead of leveraging calculations of associated features, and \textit{'w.o.F.N'} represents fine-tuning exclusively with perturbing labels, omitting feature noises.
\par
Compared with RND, the collaboration of MLLM guidance evidently facilitates the complete forgetting of target data, with over 50\% reduction of retain accuracy and 20\% decreased MIA on class rocket. It is attributed to the fine realignment effect of perturbing labels, which prevents the connections between retained feature patterns and target concepts. Without the MLLM guidance, the Fragment-Align strategy would not have functioned correctly, leading the model towards possibly the wrong tuning direction.
Meanwhile, it is demonstrated that the addition of feature noises, including a pair of positive and negative feature noise, effectively disentangles and manipulates the generalization of the forget data. This guarantees a balance between complete unlearning of forgotten data and preservation of model generalization on retained data. For unlearning on class mushroom using ViT, feature noises help increase the retain accuracy by 0.54\% while reducing forget accuracy by 10.17\%.


\begin{table}[h]
\caption{Additional results of class-wise unlearning on CIFAR-100.}\label{table:appendix class-wise CIFAR-100}
\vspace{0mm}
\centering
\begin{tabular}{l|c|cccccccc}
\toprule[1.3pt]
model                 & class                     & metric & baseline & retrain       & FT            & UNSIR         & AMNC          & SSD            & MeGU                               \\ \hline
\multirow{15}{*}{RN18} & \multirow{3}{*}{baby}     & $A_r$   & 76.54    & 75.71         & 64.64         & 71.86         & 73.53         & 76.64          & \textbf{76.70}                     \\
                      &                           & $A_f$   & 67.80    & \textbf{0.00} & \textbf{0.00} & \textbf{0.00} & \textbf{0.00} & \textbf{0.00}  & \textbf{0.00}                      \\
                      &                           & MIA    & 96.27    & 3.44          & 19.30         & 18.46         & 54.81         & \textbf{0.00}  & \textbf{0.00}                      \\ \cline{2-10} 
                      & \multirow{3}{*}{lamp}     & $A_r$   & 76.50    & 75.33         & 64.74         & 71.64         & 73.32         & \textbf{76.46} & 76.11                              \\
                      &                           & $A_f$   & 69.10    & \textbf{0.00} & \textbf{0.00} & \textbf{0.00} & \textbf{0.00} & \textbf{0.00}  & \textbf{0.00}                      \\
                      &                           & MIA    & 96.20    & 0.63          & 10.74         & 8.68          & 51.89         & \textbf{0.00}  & \textbf{0.00}                      \\ \cline{2-10} 
                      & \multirow{3}{*}{sea}      & $A_r$   & 76.35    & 73.16         & 64.00         & 71.14         & 73.45         & \textbf{75.11} & 73.36                              \\
                      &                           & $A_f$   & 83.68    & \textbf{0.00} & \textbf{0.00} & \textbf{0.00} & \textbf{0.00} & \textbf{0.00}  & \textbf{0.00}                      \\
                      &                           & MIA    & 9.42     & 4.85          & 34.41         & 36.67         & 27.66         & 1.60           & \textbf{1.20}                      \\ \cline{2-10} 
                      & \multirow{3}{*}{DINO} & $A_r$   & 76.39    & 74.79         & 63.54         & 71.63         & 73.43         & 76.39          & \textbf{76.55}                     \\
                      &                           & $A_f$   & 78.13    & \textbf{0.00} & \textbf{0.00} & \textbf{0.00} & \textbf{0.00} & \textbf{0.00}  & \textbf{0.00}                      \\
                      &                           & MIA    & 98.20    & \textbf{0.00} & 12.24         & 5.87          & 36.86         & \textbf{0.00}  & \textbf{0.00}                      \\ \cline{2-10} 
                      & \multirow{3}{*}{wolf}     & $A_r$   & 76.38    & 76.28         & 63.45         & 71.75         & 72.64         & 76.26          & \textbf{76.30} \\
                      &                           & $A_f$   & 79.51    & \textbf{0.00} & \textbf{0.00} & \textbf{0.00} & \textbf{0.00} & \textbf{0.00}  & \textbf{0.00}                      \\
                      &                           & MIA    & 97.40    & 0.49          & 13.62         & 13.84         & 43.45         & \textbf{0.00}  & \textbf{0.00}                      \\ \bottomrule[1.3pt]
\end{tabular}
\vspace{0mm}
\end{table}

\begin{table}[h]
\caption{Additional results of sub-class unlearning on CIFAR-20.}\label{table:appendix sub-class CIFAR-20}
\vspace{0mm}
\centering

\begin{tabular}{l|c|cccccccc}
\toprule[1.3pt]
model                  & subclass                    & metric & baseline & retrain        & FT    & UNSIR & AMNC       & SSD           & MeGU                      \\ \hline
\multirow{15}{*}{RN18} & \multirow{3}{*}{beetle}     & $A_r$   & 82.63    & 82.35          & 73.23 & 82.00 & 82.20          & 80.10         & \textbf{82.37}            \\
                       &                             & $A_f$   & 82.64    & 67.71          & 72.05 & 76.13 & 40.97          & 0.78 & \textbf{70.40}                     \\
                       &                             & MIA    & 88.43    & 20.26          & 58.64 & 71.00 & 9.40           & 8.81          & \textbf{6.26}             \\ \cline{2-10} 
                       & \multirow{3}{*}{snail}      & $A_r$   & 83.00    & 81.97          & 75.49 & 80.98 & 82.80          & \textbf{82.77}         & 83.10            \\
                       &                             & $A_f$   & 69.88    & 37.59          & 16.32 & 55.30 & 9.90           & 6.68 & \textbf{50.27}                     \\
                       &                             & MIA    & 84.50    & 9.47           & 16.57 & 47.22 & 10.05          & 4.80          & \textbf{4.33}             \\ \cline{2-10} 
                       & \multirow{3}{*}{whale}      & $A_r$   & 82.90    & 82.09          & 74.81 & 81.66 & \textbf{82.08}          & 82.67         & 82.74            \\
                       &                             & $A_f$   & 77.08    & 67.27          & 61.37 & 79.08 & 20.92 & 72.57         & \textbf{72.55}                     \\
                       &                             & MIA    & 85.86    & 36.47          & 54.39 & 77.80 & \textbf{3.87}  & 61.42         & 26.71                     \\ \cline{2-10} 
                       & \multirow{3}{*}{fox}        & $A_r$   & 82.96    & 82.40 & 71.50 & 82.06 & 81.98          & 79.90         & \textbf{82.32}                     \\
                       &                             & $A_f$   & 72.40    & 6.08           & 17.71 & 56.68 & \textbf{4.34}           & 0.00 & 13.12                     \\
                       &                             & MIA    & 82.46    & 5.60           & 15.66 & 35.43 & 14.61          & 21.80         & \textbf{3.70}             \\ \cline{2-10} 
                       & \multirow{3}{*}{SCP} & $A_r$   & 82.80    & 82.08          & 74.00 & 81.39 & 82.35 & 81.62         & \textbf{82.29} \\
                       &                             & $A_f$   & 90.36    & 67.53          & 61.02 & 82.12 & 5.12  & 7.46          & \textbf{69.14}                     \\
                       &                             & MIA    & 91.49    & 16.20          & 33.00 & 56.02 & 4.28           & 5.45          & \textbf{3.94}      \\ \bottomrule[1.3pt]
\end{tabular}
\vspace{0mm}
\end{table}

\subsection{Supplementary experiments}\label{sec:supplementary experiments}

In Section \ref{sec:experiments}, we demonstrate the effectiveness of MeGU in unlearning through a series of experiments, encompassing both class-wise and sub-class unlearning tasks on CIFAR-100 and CIFAR-20 datasets. To provide a more comprehensive evaluation of its performance, we extend the unlearning implementation to a larger number of classes (sub-classes).

\paragraph{Additional results of class-wise unlearning} Table \ref{table:appendix class-wise CIFAR-100} presents results of unlearning across five different classes, where \textit{DINO} denotes \textit{dinosaur}. Compared to existing methods, MeGU-unlearned models exhibit a competitive alignment effect towards the retrained model. The retain accuracy of MeGU is notably maintained, showing an increase of 0.16\% over SSD for the class \textit{dinosaur}, which can be attributed to the disentanglement of retained and forgotten concepts. Additionally, it is observed that the MIA (Membership Inference Attack) risk of MeGU-unlearned models is significantly reduced. For example, the class \textit{sea}, which presents a challenge even for the retrained model, shows a 0.4\% reduction in MIA, bottoming 1.20\% with MeGU, which is substantially lower than that of the retrained model. This further validates the security of the unlearning method proposed in our work.

\paragraph{Additional results of sub-class unlearning} Table \ref{table:appendix sub-class CIFAR-20} presents the results of sub-classes as unlearning targets, where \textit{SCP} denotes \textit{skyscraper}. While certain sub-classes, such as \textit{whale} and \textit{skyscraper}, present more challenges in the unlearning process, MeGU still outperforms existing methods by maintaining a high alignment with the accuracy of the retrained models. For sub-classes \textit{beetle}, \textit{snail}, \textit{whale}, and \textit{skyscraper}, MeGU sticks closely to the retrained models in terms of forgetting accuracy. Additionally, it significantly reduces MIA risks across all conditions. These results further demonstrate the effectiveness of our proposed method in sub-class unlearning tasks.

\end{document}